\documentclass[conference]{IEEEtran}

\usepackage{epstopdf}
\usepackage{epsfig}
\usepackage{booktabs} 
\usepackage{array}
\usepackage{lipsum}
\usepackage[]{algorithm2e}
\usepackage{tabularx}
\PassOptionsToPackage{bookmarks=false}{hyperref}
\usepackage{url}
\usepackage{graphicx}
\usepackage{booktabs}
\usepackage{float}
\usepackage{multirow}
\usepackage{multicol}
\usepackage{array}
\usepackage{tabularx}
\usepackage{bm}
\usepackage{soul}
\usepackage{tikz}
\usepackage{amssymb}
\usepackage{amsmath}
\usepackage{xfrac}
\usepackage[bottom]{footmisc}
\usepackage{fixltx2e}
\usepackage{etoolbox}

\IEEEtriggeratref{1}

\begin{document}
\title{Prive-HD: Privacy-Preserved Hyperdimensional Computing}

\author{Behnam Khaleghi, Mohsen Imani, Tajana Rosing\\
\IEEEauthorblockA{CSE Department, UC San Diego, La Jolla, CA 92093, USA\\
    \{bkhaleghi, moimani, tajana\}@ucsd.edu}\vspace{-0.0cm}
}

\maketitle


\begin{abstract}
The privacy of data is a major challenge in machine learning as a {trained model} may expose sensitive information of the enclosed dataset.
Besides, the limited computation capability and capacity of edge devices have made cloud-hosted inference inevitable.
Sending private information to remote servers makes the privacy of {inference} also vulnerable because of susceptible communication channels or even untrustworthy hosts.
In this paper, we target privacy-preserving training and inference of brain-inspired Hyperdimensional (HD) computing, a new learning algorithm that is gaining traction due to its light-weight computation and robustness particularly appealing for edge devices with tight constraints.
Indeed, despite its promising attributes, HD computing has virtually no privacy due to its reversible computation.
We present an accuracy-privacy trade-off method through meticulous quantization and pruning of hypervectors, the building blocks of HD, to realize a {differentially private} model as well as to obfuscate the information sent for cloud-hosted inference.
Finally, we show how the proposed techniques can be also leveraged for efficient hardware implementation.

\end{abstract}

\section{Introduction}\label{sec:intro}
The efficacy of machine learning solutions in performing various tasks has made them ubiquitous in different application domains. 
The performance of these models is proportional to the size of the training dataset.
Thus, machine learning models utilize copious proprietary and/or crowdsourced data, e.g., medical images.
In this sense, different privacy concerns arise.
The first issue is with model exposure \cite{abadi2016deep}.
Obscurity is not considered a guaranteed approach for privacy, especially parameters of a model (e.g., weights in the context of neural networks) that might be leaked through inspection.
Therefore, in the presence of an adversary with full knowledge of the trained model parameters, the model should not reveal the information of constituting records.

Second, the increasing complexity of machine learning models, on the one hand, and the limited computation and capacity of edge devices, especially in the IoT domain with extreme constraints, on the other hand, have made offloading computation to the cloud indispensable \cite{teerapittayanon2017distributed, li2018learning}.
%
An immediate drawback of cloud-based inference is compromising client data privacy.
The communication channel is not only susceptible to attacks, but an untrusted cloud itself may also expose the data to third-party agencies or exploit it for its benefits.
Therefore, transferring the least amount of information while achieving maximal accuracy is of utmost importance.
A traditional approach to deal with such privacy concerns is employing secure multi-party computation that leverages homomorphic encryption whereby the device encrypts the data, and the host performs computation on the ciphertext \cite{gilad2016cryptonets}.
These techniques, however, impose a prohibitive computation cost on edge devices.

Previous work on machine learning, particularly deep neural networks, have come up with generally two approaches to preserve the privacy of training (model) or inference.
For privacy-preserving \textit{training}, the well-known concept of \textit{differential privacy} is incorporated in the training \cite{dwork2006calibrating, mcsherry2009differentially}.
Differential privacy, often known as the standard notation of guaranteed privacy, aims to apply a carefully chosen noise distribution to make the response of a query (in our concept, the model being trained on a dataset) over a database randomized enough so the singular records remain indistinguishable whilst the query result is fairly accurate.
Perturbation of partially processed information, e.g., the output of the convolution layer in neural networks, before offloading to a remote server is another trend of privacy-preserving studies \cite{wang2018not, osia2020hybrid, mireshghallah2020shredder} that target the \textit{inference} privacy.
Essentially, it degrades the {mutual information} of the conveyed data. 
This approach degrades the prediction accuracy and requires (re)-training the neural network to compensate the injected noise \cite{wang2018not} or analogously learning the parameters of a noise that can be tolerated by the network \cite{mireshghallah2020shredder, mireshghallah2020principled}, which are not always feasible, e.g., when the model is inaccessible.

In this paper, for the first time, we scrutinize Hyperdimensional (HD) computing from a privacy perspective.
HD is a novel efficient learning paradigm that imitates the brain functionality in cognitive tasks, in the sense that the human brain computes with patterns of neural activity rather than scalar values \cite{kanerva2009hyperdimensional, schmuck2019hardware, mitrokhin2019learning, neubertintroduction}.
These patterns and underlying computations can be realized by points and light-weight operations in a hyperdimensional space, i.e., by hypervectors of $\sim$10,000 dimensions.
Similar to other statistical {mechanisms}, the privacy of HD might be preserved by noise injection, where formally the granted privacy budget is directly proportional to the amount of the introduced noise and \textit{indirectly} to the \textit{sensitivity} of mechanism.
Nonetheless, as a query hypervector (HD's raw output) has thousands of $w$-bits dimensions, the sensitivity of the HD model can be extremely large, which requires a tremendous amount of noise to guarantee differential privacy, which significantly reduces accuracy.
Similarly, the magnitude of each output dimension is large (each up to $2^w$), so is the intensity of the required noise to disguise the transferring information for inference.
Therefore, we require more prudent approaches to augment HD with differentially private training as well as blurring the information of offloaded inference.

Our main contributions are as follows.
We show the privacy breach of HD and introduce different techniques including well-devised hypervector (query and/or class) quantization and dimension pruning to reduce the sensitivity, and consequently, the required noise to achieve a differentially private HD model.
We also target inference privacy by showing how quantizing the query hypervector, during inference, can achieve good prediction accuracy as well as multifaceted power efficiency while significantly degrading the Peak Signal-to-Noise Ratio (PSNR) of reconstructed inputs (i.e., diminishing useful transferred information).
Finally, we propose an approximate hardware implementation that benefits from the aforementioned innovations for further performance and power efficiency. 

\section{Preliminary}\label{sec:pre}

\subsection{Hyperdimensional Computing}
\textbf{Encoding} is the first and major operation involved in both training and inference of HD.
Assume that an input vector (an image, voice, etc.) comprises $\mathcal{D}_{iv}$ dimensions (elements or features).
Thus, each input $\vec{\mathcal{V}}_{iv}$ can be represented as \eqref{eq:rep1}.
`$v_i$'s are elements of the input, where each feature $v_i$ takes value among ${f}_0$ to ${f}_{\ell_{iv}-1}$.
In a black and white image, there are only two feature levels ($\ell_{iv}=2$), and ${f}_0=0$, and ${f}_1 = 1$.

\begin{equation}\label{eq:rep1}
\begin{split}
\vec{\mathcal{V}}_{iv} = \langle v_0, v_1, \cdots, v_{\mathcal{D}_{iv}-1} \rangle \\
|v_i| \in \mathcal{F} = \{{f}_0, {f}_1, \cdots {f}_{\ell_{iv}-1}\}
\end{split}
\end{equation}

Varied HD encoding techniques with different accuracy-performance trade-off have been proposed \cite{kanerva2009hyperdimensional, imani2019framework}.
Equation \eqref{eq:enc} shows analogous encodings that yield accuracies similar to or better than the state of the art \cite{imani2019framework}.

\begin{subequations}\label{eq:enc}
\begin{equation}\label{eq:enca}
\vec{\mathcal{H}} = \sum_{k=0}^{\mathcal{D}_{iv}-1} |v_k|_{\in \mathcal{F}} \cdot \vec{\mathcal{B}}_{k}
\end{equation}
\begin{equation}\label{eq:encb}
\vec{\mathcal{H}} = \sum_{k=0}^{\mathcal{D}_{iv}-1} \vec{\mathcal{L}}_{v_k} \cdot \vec{\mathcal{B}}_{k}
\end{equation}
\end{subequations}

\noindent `$\vec{\mathcal{B}}_{k}$'s are randomly chosen hence orthogonal bipolar base hypervectors of dimension $\mathcal{D}_{hv} \simeq 10^4$ to retain the spatial or temporal location of features in an input.
That is, $\vec{\mathcal{B}}_{k} \in \{-1, +1\}^{\mathcal{D}_{hv}}$ and $\delta(\vec{\mathcal{B}}_{k_1}, \vec{\mathcal{B}}_{k_2}) \simeq 0$, where $\delta$ denotes the cosine similarity: $\delta(\vec{\mathcal{B}}_{k_1}, \vec{\mathcal{B}}_{k_2}) = \frac{\vec{\mathcal{B}}_{k_1} \cdot \vec{\mathcal{B}}_{k_2}}{\mathbin{\parallel} \vec{\mathcal{B}}_{k_1} \mathbin{\parallel} \cdot \mathbin{\parallel} \vec{\mathcal{B}}_{k_2} \mathbin{\parallel}}$.
Evidently, there are $\mathcal{D}_{iv}$ fixed base/location hypervectors for an input (one per feature).
The only difference of the encodings in \eqref{eq:enca} and \eqref{eq:encb} is that in \eqref{eq:enca} the scalar value of each input feature $v_k$ (mapped/quantized to nearest $f$ in $\mathcal{F}$) is directly multiplied in the corresponding base hypervector $\vec{\mathcal{B}}_{k}$.
However, in \eqref{eq:encb}, there is a level hypervector of the same length ($\mathcal{D}_{hv}$) associated with different feature values.
Thus, for $k^{th}$ feature of the input, instead of multiplying $f_{|v_k|} \simeq |v_k|$ by location vector $\vec{\mathcal{B}}_{k}$, the associated hypervector $\vec{\mathcal{L}}_{v_k}$ performs a dot-product with $\vec{\mathcal{B}}_{k}$.
As both vectors are binary, the dot-product reduces to dimension-wise XNOR operations.
To maintain the \textit{closeness} in features (to demonstrate closeness in original feature values), $\vec{\mathcal{L}}_0$ and $\vec{\mathcal{L}}_{\ell_{iv}-1}$ are entirely orthogonal, and each $\vec{\mathcal{L}}_{k+1}$ is obtained by flipping randomly chosen $\frac{\mathcal{D}_{hv}}{2 \cdot \ell_{iv}}$ bits of $\vec{\mathcal{L}}_{k}$.

\textbf{Training} of HD is simple.
After generating each encoding hypervector $\vec{\mathcal{H}}^l$ of inputs belonging to class/label $l$, the class hypervector $\vec{\mathcal{C}^l}$ can be obtained by bundling (adding) all $\vec{\mathcal{H}}^l$s.
Assuming there are $\mathcal{J}$ inputs having label $l$:

\begin{equation}\label{eq:class}
\vec{\mathcal{C}^l} = \sum_{j}^{\mathcal{J}}{\vec{\mathcal{H}^l_{j}}}
\end{equation}

\textbf{Inference} of HD has a two-step procedure.
The first step encodes the input (similar to encoding during training) to produce a query hypervector $\vec{\mathcal{H}}$.
Thereafter, the similarity ($\delta$) of $\vec{\mathcal{H}}$ and all class hypervectors are obtained to find out the class with highest similarity:

\begin{equation}\label{eq:cosine}
\delta(\vec{\mathcal{H}}, \vec{\mathcal{C}^l}) = \frac{\vec{\mathcal{H}} \cdot \vec{\mathcal{C}^l}}{\mathbin{\parallel} \vec{\mathcal{H}} \mathbin{\parallel} \cdot \mathbin{\parallel} \vec{\mathcal{C}^l} \mathbin{\parallel}} = \frac{\sum_{k=0}^{\mathcal{D}_{hv}-1}{h_k \cdot c^l_k}}{\sqrt{\sum_{k=0}^{\mathcal{D}_{hv}-1}{h^2_k}} \cdot \sqrt{\sum_{k=0}^{\mathcal{D}_{hv}-1}{{c^l}^2_k}}}
\end{equation}

\noindent Note that $\sqrt{\sum_{k=0}^{\mathcal{D}_{hv}-1}{h^2_k}}$ is a repeating factor when comparing with all classes, so can be discarded.
The $\sqrt{\sum_{k=0}^{\mathcal{D}_{hv}-1}{{c^l}^2_k}}$ factor is also constant for a classes, so only needs to be calculated once.

\textbf{Retraining} can boost the accuracy of the HD model by discarding the mispredicted queries from corresponding mispredicted classes, and adding them to the right class.
Retraining examines if the model correctly returns the label $l$ for an encoded query $\vec{\mathcal{H}}$.
If the model mispredicts it as label $l'$, the model updates as follows.

\begin{equation}\label{eq:update}
\begin{split}
\vec{\mathcal{C}^l} = \vec{\mathcal{C}^l} + \mathcal{\vec{H}} \\
\vec{\mathcal{C}^{l'}} = \vec{\mathcal{C}^{l'}} - \mathcal{\vec{H}}
\end{split}
\end{equation}

\subsection{Differential Privacy}

Differential privacy targets the indistinguishability of a mechanism (or algorithm), meaning whether observing the output of an algorithm, i.e., computations' result, may disclose the computed data.
Consider the classical example of a sum query $f(n) = \sum_{1}^{n} g(x_i)$ over a database with $x_i$s being the first to $n^{th}$ rows, and $g(x_i) \in \{0, 1\}$, i.e., the value of each record is either $0$ or $1$.
Although the function $f$ does not reveal the value of an arbitrary record $m$, it can be readily obtained by two requests as $f(m) - f(m-1)$.
Speaking formally, a randomized algorithm $\mathcal{M}$ is $\varepsilon$-indistinguishable or $\varepsilon$-differentially private if for any inputs $\mathcal{D}_1$ and $\mathcal{D}_2$ that differ in one entry (a.k.a \textit{adjacent} inputs) and any output $\mathcal{S}$ of $\mathcal{M}$, the following holds:

\begin{equation}\label{eq:dif1}
Pr[\mathcal{M}(\mathcal{D}_1) \in S] \leq e^\varepsilon \cdot Pr[\mathcal{M}(\mathcal{D}_2) \in S]
\end{equation}

This definition guarantees that observing $\mathcal{D}_1$ instead of $\mathcal{D}_2$ scales up the probability of any event by no more than $e^{\varepsilon}$.
Evidently, smaller values of non-negative $\varepsilon$ provide stronger guaranteed privacy.
Dwork et al. have shown that $\varepsilon$-differential privacy can be ensured by adding a Laplace noise of scale Lap($\frac{\Delta f}{\varepsilon}$) to the output of algorithm \cite{dwork2006calibrating}.
$\Delta f$, defined as $\ell_1$ norm in Equation \eqref{eq:sen}, denotes the sensitivity of the algorithm which represents the amount of change in a mechanism's output by changing one of its arguments, e.g., inclusion/exclusion of an input in training.

\begin{equation}\label{eq:sen}
\Delta f = \mathbin{\parallel} f(\mathcal{D}_1) - f(\mathcal{D}_2) \mathbin{\parallel}_1
\end{equation}

Dwork et al. have also introduced a more amiable $\delta$-approximate $\varepsilon$-indistinguishable privacy guarantee, which allows the $\varepsilon$-privacy to be broken by a probability of $\delta$ \cite{dwork2006our}.

\begin{equation}\label{eq:dif2}
\mathcal{M}(\mathcal{D}) = f(\mathcal{D}) + \mathcal{G}(0, \Delta f^2 \sigma^2)
\end{equation}

$\mathcal{G}(0, \Delta f^2  \sigma^2)$ is Gaussian noise with mean zero and standard deviation of $\Delta f \cdot \sigma$.
Both $f$ and $\mathcal{G}$ are $D_{hv} |\mathcal{C}|$ dimensions, i.e., $|\mathcal{C}|$ output class hypervectors of $D_{hv}$ dimensions.
Here, $\Delta f = \mathbin{\parallel} f(\mathcal{D}_1) - f(\mathcal{D}_2) \mathbin{\parallel}_2$ is $\bm{\ell_2}$ \textbf{norm}, which relaxes the amount of additive noise.
$f$ meets $(\varepsilon, \delta)$-privacy if $\delta \geq \frac{4}{5} e^{-\frac{(\sigma \varepsilon)^2}{2}}$ \cite{abadi2016deep}.
Achieving small $\varepsilon$ for a given $\delta$ needs larger $\sigma$, which by \eqref{eq:dif2} translates to larger noise.



\section{Proposed Method: Prive-HD}

\subsection{Privacy Breach of HD} \label{subsec:breach}

\begin{figure}[t]
  \centering
  \includegraphics[width=0.72\columnwidth]{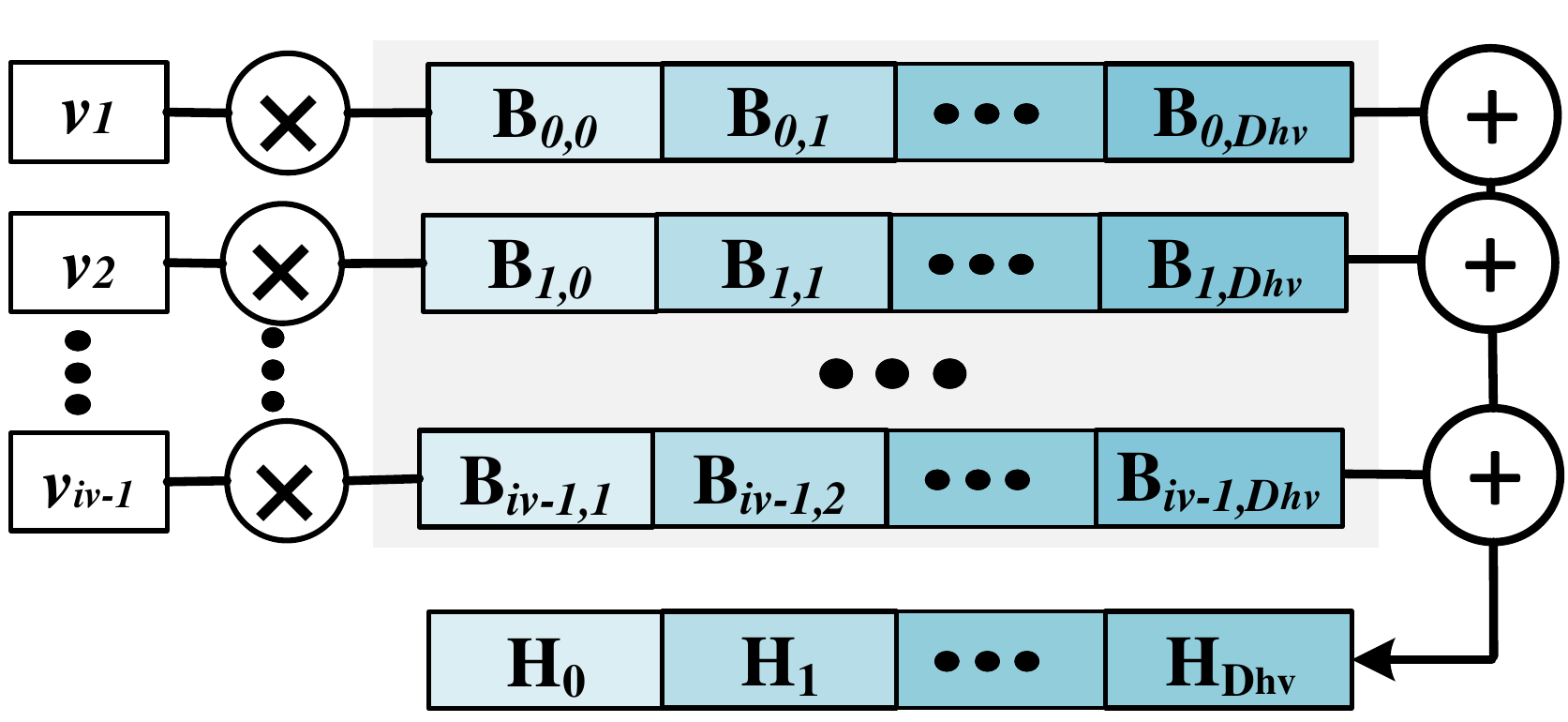} \vspace{-0.0cm}
  \caption{Encoding presented in Equation \eqref{eq:enca}.}\label{fig:encoding}  \vspace{-0.0cm}
\end{figure}

In contrast to the deep neural networks that comprise non-linear operations that somewhat cover up the details of raw input, HD operations are fairly reversible, leaving it zero privacy.
That is, the input can be reconstructed from the encoded hypervector.
Consider the encoding of Equation \eqref{eq:enca}, which is also illustrated by Fig. \ref{fig:encoding}.
Multiplying each side of the equation to hypevector $\vec{\mathcal{B}}_{0}$, for each dimension $j$ gives:

\begin{equation}\label{eq:reverse1}
\begin{split}
\vec{\mathcal{H}}_{j} \cdot \mathcal{B}_{0,j} = \sum_{k=0}^{\mathcal{D}_{iv}-1}{(|v_k| \cdot \mathcal{B}_{k,j} ) \cdot \mathcal{B}_{0,j} } = \qquad \qquad \\
|v_0| \cdot \mathcal{B}_{0,j}^2 + \sum_{k=1}^{\mathcal{D}_{iv}-1}{|v_k| \mathcal{B}_{k,j} \mathcal{B}_{0,j} } = |v_0| + \sum_{k=1}^{\mathcal{D}_{iv}-1}{|v_k| \mathcal{B}_{k,j} \mathcal{B}_{0,j} }
\end{split}
\end{equation}

\noindent  $\mathcal{B}_{0,j} \in \{-1, +1\}$, so $\mathcal{B}_{0,j}^2 = 1$.
Summing all dimensions together yields:

\begin{equation}\label{eq:reverse2}
\sum_{j=0}^{\mathcal{D}_{hv}-1}{\vec{\mathcal{H}}_{j} \cdot \mathcal{B}_{0,j}} = \mathcal{D}_{hv} \cdot |v_0| + \sum_{k=1}^{\mathcal{D}_{iv}-1}{\big(|v_k| \sum_{j=0}^{\mathcal{D}_{hv}-1}{\mathcal{B}_{k,j} \cdot \mathcal{B}_{0,j}}\big)}
\end{equation}

\begin{figure}[t]
  \centering
  \includegraphics[width=0.6\columnwidth]{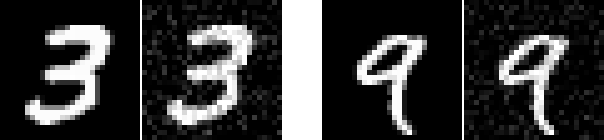} \vspace{-0.0cm}
  \caption{Original and retrieved handwritten digits.}\label{fig:decoding} \vspace{-0.0cm}
\end{figure}

As the base hypervectors are orthogonal and especially $\mathcal{D}_{hv}$ is large, $\sum_{j=0}^{\mathcal{D}_{hv}-1}{\mathcal{B}_{k,j} \cdot \mathcal{B}_{0,j}} \simeq 0$ in the right side of Equation \eqref{eq:reverse2}.
It means that every feature $|v_m|$ can be retrieved back by $|v_m| = \frac{\vec{\mathcal{H}} \cdot \vec{\mathcal{B}_{m}}}{\mathcal{D}_{hv}}$.
Note that without lack of generality we assumed $|v_m| = f_{v_m}$, i.e., features are not normalized or quantized.
Indeed, we are retrieving the \textit{features} (`$f_i$'s), that might or might not be the exact raw elements.
Also, although we showed the reversibility of the encoding in \eqref{eq:enca}, it can easily be adjusted to the other HD encodings.
Fig. \ref{fig:decoding} shows the reconstructed inputs of MNIST samples by using Equation \eqref{eq:reverse2} to achieve each of $28 \times 28$ pixels, one by one.

That being said, the encoded hypervector $\vec{\mathcal{H}}$ sent for cloud-hosted inference can be inspected to reconstruct the original input.
This reversibility also breaches the privacy of the HD model.
Consider that, according to the definition of differential privacy, two datasets $\mathcal{D}_1$ and $\mathcal{D}_2$ differ by one input.
If we subtract all class hypervectors of the models trained over $\mathcal{D}_1$ and $\mathcal{D}_2$, 
the result (difference) will exactly be the encoded vector of the missing input (remember from Equation \eqref{eq:class} that class hypervectors are simply created by adding encoded hypervectors of associated inputs).
The encoded hypervector, hence, can be decoded back to obtain the missing input.

\subsection{Differentially Private HD Training}
Let $f_{\mathcal{D}_1}$ and $f_{\mathcal{D}_2}$ be models trained with encoding of Equation \eqref{eq:enca} over datasets that differ in a single datum (input) present in $\mathcal{D}_2$ but not in $\mathcal{D}_1$.
The outputs (i.e., class hypervectors) of $f(\mathcal{D}_1)$ and $f(\mathcal{D}_2)$ thus differ in inclusion of a single $\mathcal{D}_{hv}-$dimension encoded vector that misses from a particular class of $f(\mathcal{D}_1)$.
The other class hypervectors will be the same.
Each bipolar hypervector $\vec{\mathcal{L}}_{v_k} \cdot \vec{\mathcal{B}}_{k}$ (see Equation \eqref{eq:enc} or Fig. \ref{fig:encoding}) constituting the encoding $\vec{\mathcal{H}}$ is random and identically distributed, hence according to the central limit theorem $\vec{\mathcal{H}}$ is approximately normally distributed with $\mu=0$ and $\sigma^2=\mathcal{D}_{iv}$, i.e., the number of vectors building $\vec{\mathcal{H}}$.
In ${\ell_1}$ norm, however, the absolute value of the encoded $\vec{\mathcal{H}}$ matters.
Since $\vec{\mathcal{H}}$ has normal distribution, mean of the corresponding folded (absolute) distribution is:

\begin{equation}\label{eq:enc_param_mean}
\mu_{|\vec{\mathcal{H}}|} = \sigma \sqrt{\frac{2}{\pi}}  e^{-\frac{\mu^2}{2\sigma^2}} + \mu (1 - \Phi(-\frac{\mu}{\sigma})) = \sqrt{\frac{2\mathcal{D}_{iv}}{\pi}} 
\end{equation}
%

The ${\ell_1}$ sensitivity will therefore be $\Delta f = \mathbin{\parallel} \vec{\mathcal{H}} \mathbin{\parallel}_1 = \sqrt{\frac{2\mathcal{D}_{iv}}{\pi}} \cdot \mathcal{D}_{hv}$.
For the ${\ell_2}$ sensitivity we indeed deal with a squared Gaussian (chi-squared) distribution with freedom degree of one, thus:

\begin{equation}\label{eq:enc_param_mean2}
\Delta f = \mathbin{\parallel} \vec{\mathcal{H}} \mathbin{\parallel}_2 = \sqrt{\mathcal{D}_{hv} \cdot \mu'} = \sqrt{\mathcal{D}_{hv} \cdot \sigma^2} = \sqrt{\mathcal{D}_{hv} \cdot \mathcal{D}_{iv}}
\end{equation}

\noindent Note that the mean of the chi-squared distribution ($\mu'$) is equal to the variance ($\sigma^2$) of the original distribution of $\vec{\mathcal{H}}$.
Both Equation \eqref{eq:enc_param_mean} and \eqref{eq:enc_param_mean2} imply a large noise to guarantee privacy.
For instance, for a modest 200-features input ($\mathcal{D}_{iv}=200$) the $\ell_2$ sensitivity is $10^3\sqrt{2}$ while a proportional noise will annihilate the model accuracy.
In the following, we articulate the proposed techniques to shrink the variance of the required noise.
In the rest of the paper, we only target Gaussian noise, i.e., $(\varepsilon, \delta)-$privacy, since in our case it needs a weaker noise.


\subsubsection{\textbf{Model Pruning}}
An immediate observation from Equation \eqref{eq:enc_param_mean2} is to reduce the number of hypervectors dimension, $\mathcal{D}_{hv}$ to mollify the sensitivity, hence, the required noise.
Not all the dimensions of a class hypervector have the same impact on prediction.
Remember, from Equation \eqref{eq:cosine}, that prediction is realized by a normalized dot-product between the encoded query and class hypervectors.
Intuitively, we may prune out the close-to-zero class elements as their element-wise multiplication with query elements leads to less-effectual results.
Notice that this concept (i.e., discarding a major portion of the weights without significant accuracy loss) does not readily hold for deep neural networks as the impact of those small weights might be amplified by large activations of previous layers.
In HD, however, information is uniformly distributed over the dimensions of the query hypervector, so overlooking some of the query's information (the dimensions corresponding to discarded \textit{less-effectual} dimensions of class hypervectors) should not cause unbearable accuracy loss.

\begin{figure}[t]
  \centering
  \includegraphics[width=0.50\textwidth]{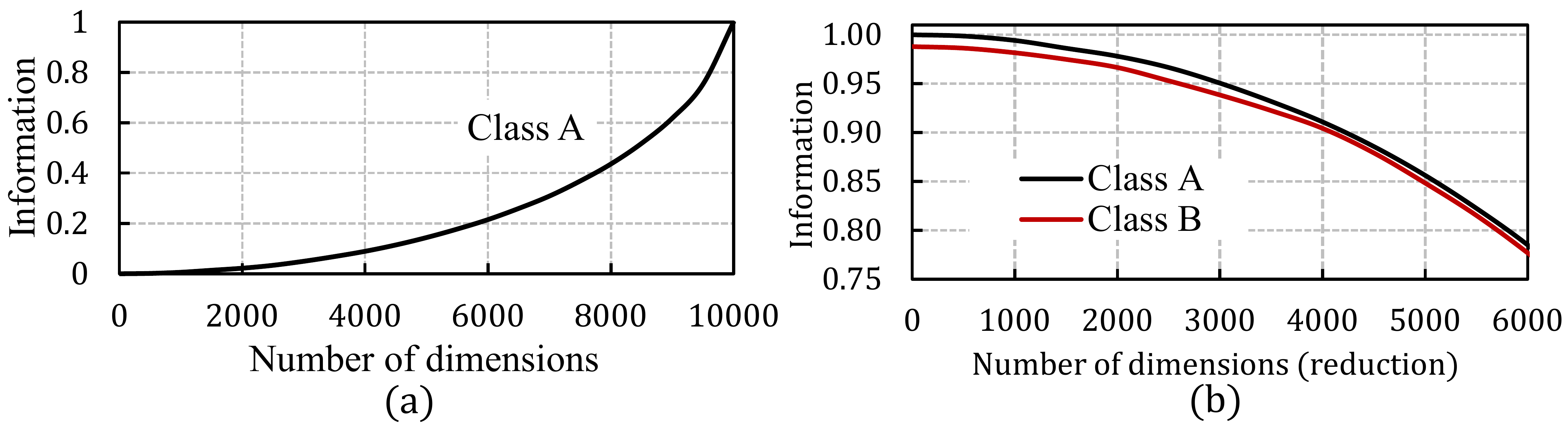} \vspace{-0.0cm}
  \caption{Impact of increasing (left) and reducing (right) \textit{effectual} dimensions.}\label{fig:dimension} \vspace{-0.0cm}
\end{figure}

We demonstrate the model pruning as an example in Fig. \ref{fig:dimension} (that belongs to a speech recognition dataset).
In Fig. \ref{fig:dimension}(a), after training the model, we remove all dimensions of a certain class hypervector.
Then we increasingly add (return) its dimensions starting from the \textit{less-effectual} dimensions.
That is, we first restore the dimensions with (absolute) values close to zero.
Then we perform a similarity check (i.e., prediction of a certain query hypervector via normalized dot-product) to figure out what portion of the original dot-product value is retrieved.
As it can be seen in the same figure, the first 6,000 close-to-zero dimensions only retrieve 20\% of the information required for a fully confident prediction.
This is because of the uniform distribution of information in the encoded query hypervector: the pruned dimensions do not correspond to vital information of queries.
Fig. \ref{fig:dimension}(b) further clarifies our observation.
Pruning the less-effectual dimensions slightly reduces the prediction information of both class A (correct class, with an initial total of 1.0) and class B (incorrect class).
As more effectual dimensions of the classes are pruned, the slope of information loss plunges.
It is worthy of note that in this example the ranks of classes A and B have been retained.

\begin{figure}[t]
  \centering
  \includegraphics[width=0.4\textwidth, height=2.6cm]{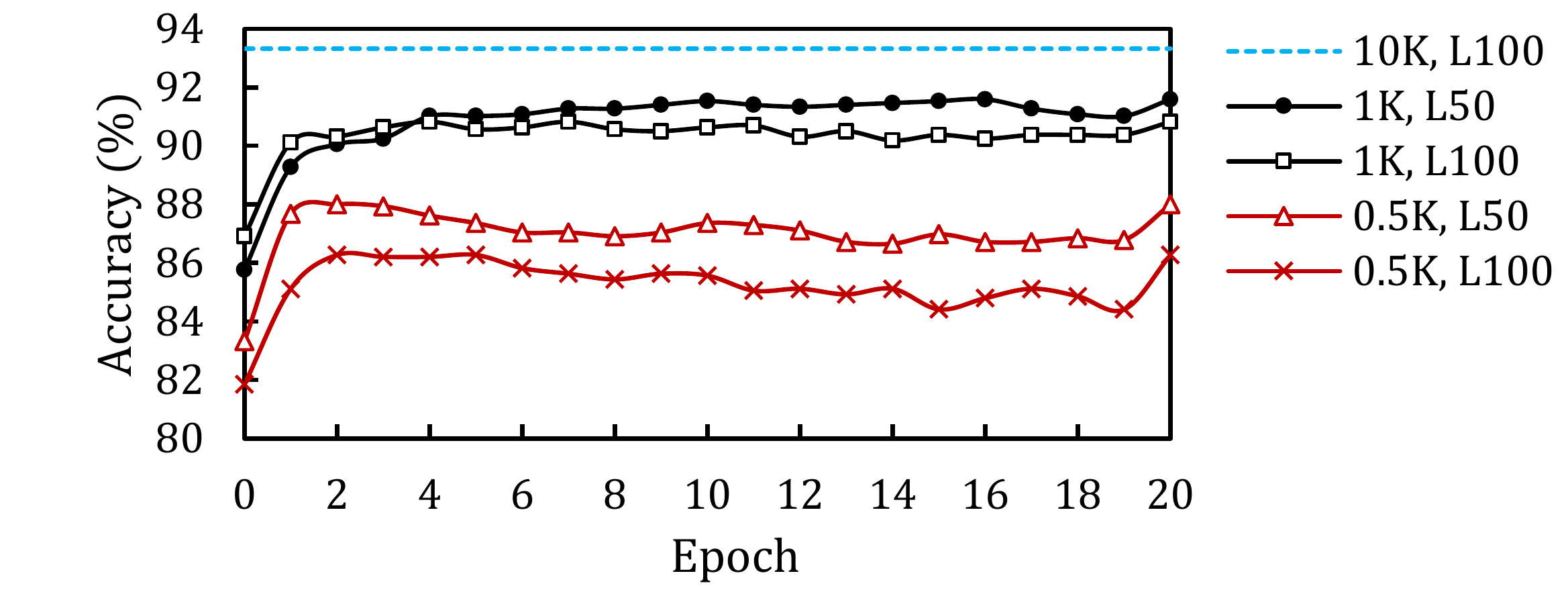} \vspace{-0.0cm}
  \caption{Retraining to recover accuracy loss.}\label{fig:accuracy} \vspace{-0.0cm}
\end{figure}

We augment the model pruning by retraining explained in Equation \eqref{eq:update} to partially recover the information of the pruned dimensions in the remaining ones.
For this, we first nullify $s\%$ of the close-to-zero dimensions of the trained model, which perpetually remain zero.
Therefore, during the encoding of query hypervectors, we do not anymore need to obtain the corresponding indexes of queries (note that operations are dimension-wise), \textit{which translates to reduced sensitivity.}
Thereafter, we repeatedly iterate over the training dataset and apply Equation \eqref{eq:update} to update the classes involved in mispredictions.
Fig. \ref{fig:accuracy} shows 1-2 iteration(s) is sufficient to achieve the maximum accuracy (the last iteration simple shows the maximum of previous ones).
In lower dimension, decreasing the number of levels ($\ell_{iv}$ in Equation \eqref{eq:rep1}, denoted by L in the legend), achieves slightly higher accuracy as hypervectors lose the capacity to embrace fine-grained details.

\subsubsection{\textbf{Encoding Quantization}}
Previous work on HD computing have introduced the concept of model quantization for compression and energy efficiency, where both encoding and class hypervectors are quantized at the cost of significant accuracy loss \cite{salamat2019f5}.
We, however, only target quantizing the encoding hypervectors since the sensitivity is merely determined by the $\ell_2$ norm of encoding.
Equation \eqref{eq:quant} shows the 1-bit quantization of encoding in \eqref{eq:enca}.
The original scalar-vector product, as well as the accumulation, is performed in full-precision, and only the final hypervector is quantized.
The resultant class hypervectors will also be \textit{non-binary} (albeit with reduced dimension values).

\begin{equation}\label{eq:quant}
\vec{\mathcal{H}}_{q1} = \text{sign}\big(\sum_{k=0}^{\mathcal{D}_{iv}-1} |v_k|_{\in \mathcal{F}} \cdot \vec{\mathcal{B}}_{k}\big)
\end{equation}

\begin{figure}[!t]
  \centering
  \includegraphics[width=0.5\textwidth, height=2.9cm]{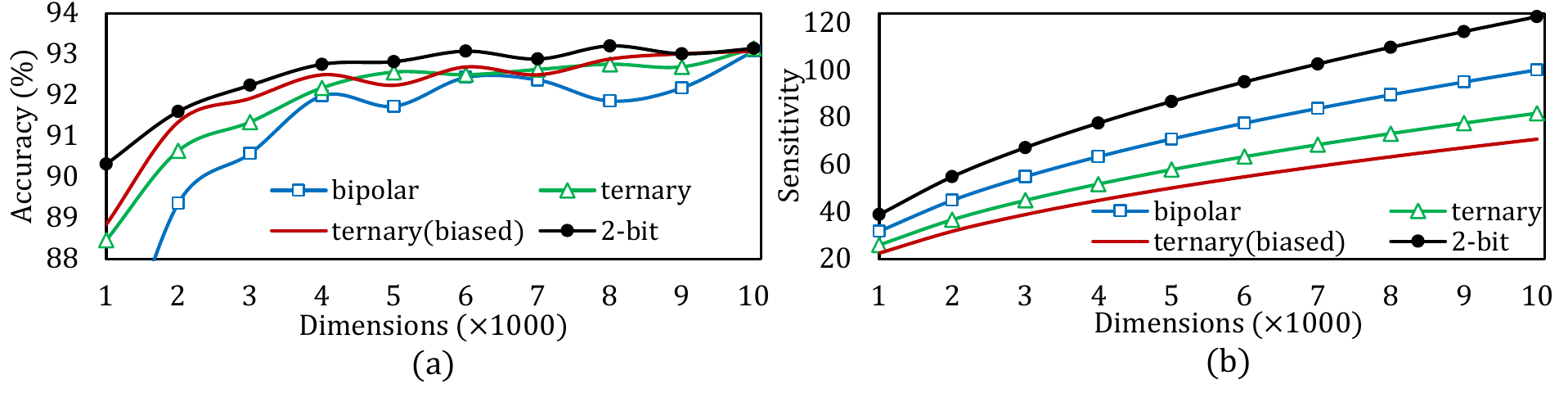}\vspace{-0.0cm}\caption{Accuracy-sensitivity trade-off of encoding quantization.}\label{fig:quant-sen}\vspace{-0.0cm}
  \end{figure}
  
Fig. \ref{fig:quant-sen} shows the impact of quantizing the encoded hypervectors on the accuracy and the sensitivity of the same speech recognition dataset trained with such encoding.
In 10,000 dimensions, the bipolar (i.e., $\pm 1$ or sign) quantization achieves 93.1\% accuracy while it is 88.1\% in previous work \cite{salamat2019f5}.
This improvement comes from the fact that we do not quantize the class hypervectors.
We then leveraged the aforementioned pruning approach to simultaneously employ quantization and pruning, as demonstrated in Fig. \ref{fig:quant-sen}(a).
In $\mathcal{D}_{hv}=1000$, the 2-bit quantization ($\{-2, \pm 1, 0\}$) achieves 90.3\% accuracy, which is only 3\% below the full-precision full-dimension baseline.
It should note be noted that the small oscillations in specific dimensions, e.g., lower accuracy in 5,000 dimensions compared to 4,000 dimensions in bipolar quantization, are due to randomness of the initial hypervectors and non-orthogonality that show up in smaller space.

Fig. \ref{fig:quant-sen}(b) shows the sensitivities of the corresponding models.
After quantizing, the number of features, $\mathcal{D}_{iv}$ (see Equation \eqref{eq:enc_param_mean2}), does not matter anymore.
The sensitivity of a quantized model can be formulated as follows.

\begin{equation}\label{eq:sens2}
\Delta f = \mathbin{\parallel} \vec{\mathcal{H}} \mathbin{\parallel}_2 = (\sum_{k \in |q|}{p_k \cdot \mathcal{D}_{hv} \cdot k^2})^{1/2}
\end{equation}

\noindent $p_k$ shows the probability of $k$ (e.g., $\pm 1$) in the quantized encoded hypervector, so $p_k \cdot \mathcal{D}_{hv}$ is the total occurrence of $k$ quantized encoded hypervector.
The rest is simply the definition of $\ell_2$ norm.
As hypervectors are randomly generated and i.i.d, the distribution of $k \in |q|$ is uniform.
That is, in the bipolar quantization, roughly $\mathcal{D}_{hv}/2$ of encoded dimensions are $1$ (or $-1$).
We therefore also exploited a \textit{biased} quantization to give more weight for $p_0$ in the ternary quantization, dubbed as `ternary (biased)' in Fig. \ref{fig:quant-sen}(b).
Essentially the biased quantization assigns a quantization threshold to conform to $p_{-1} = p_1 = \frac{1}{4}$, while $p_0 = \frac{1}{2}$.
This reduces the sensitivity by a factor of $\frac{\sqrt{\frac{\mathcal{D}_{hv}}{4}+\frac{\mathcal{D}_{hv}}{4}}}{\sqrt{\frac{\mathcal{D}_{hv}}{3}+\frac{\mathcal{D}_{hv}}{3}}} = 0.87\times$.
Combining quantizatoin and pruning, we could shrink the $\ell_2$ sensitivity to $\Delta f=22.3$, which originally was $\sqrt{10^4 \cdot 617} = 2484$ for the speech recognition with 617-features inputs.
In Section \ref{sec:res} we will examine the impact of adding such noise on the model accuracy for varied privacy budgets.

\subsection{Inference Privacy}\label{subsec:inference}
Building upon the multi-layer structure of ML, IoT devices mostly rely on performing primary (e.g., feature extraction) computations on the edge (or edge server) and offload the decision-making final layers to the cloud \cite{teerapittayanon2017distributed, li2018learning}.
To tackle the privacy challenges of offloaded inference, previous work on DNN-based inference generally inject noise on the offloaded computation.
This necessitates either to retrain the model to tolerate the injected noise distribution \cite{wang2018not}, or analogously, learn the parameters of a noise that maximally perturbs the information with preferably small impact on the accuracy \cite{mireshghallah2020shredder, mireshghallah2020principled}.

In Section \ref{subsec:breach} we demonstrated how the original feature vector can be reconstructed from the encoding hypervectors.
Inspired by the encoding quantization technique explained in the previous section, we introduce a turnkey technique to obfuscate the conveyed information without manipulating or even accessing the model.
Indeed, we observed that quantizing down to 1-bit (bipolar) even in the presence of model pruning could yield acceptable accuracy.
As shown in Fig. \ref{fig:quant-sen}(a), 1-bit quantization only incurred 0.25\% accuracy loss.
These models, however, were trained by accumulating quantized encoding hypervectors.
Intuitively, we expect that performing inference with quantized query hypervectors but on full-precision classes (class hypervectors generated by non-quantized encoding hypervectors) should give the same or better accuracy as quantizing is nothing but degrading the information.
In other words, in the previous case, we deal with checking the similarity of a degraded query with classes built up also from degraded information, but now we check the similarity of a degraded query with information-rich classes.

\begin{figure}[t]
  \centering
  \includegraphics[width=0.45\textwidth]{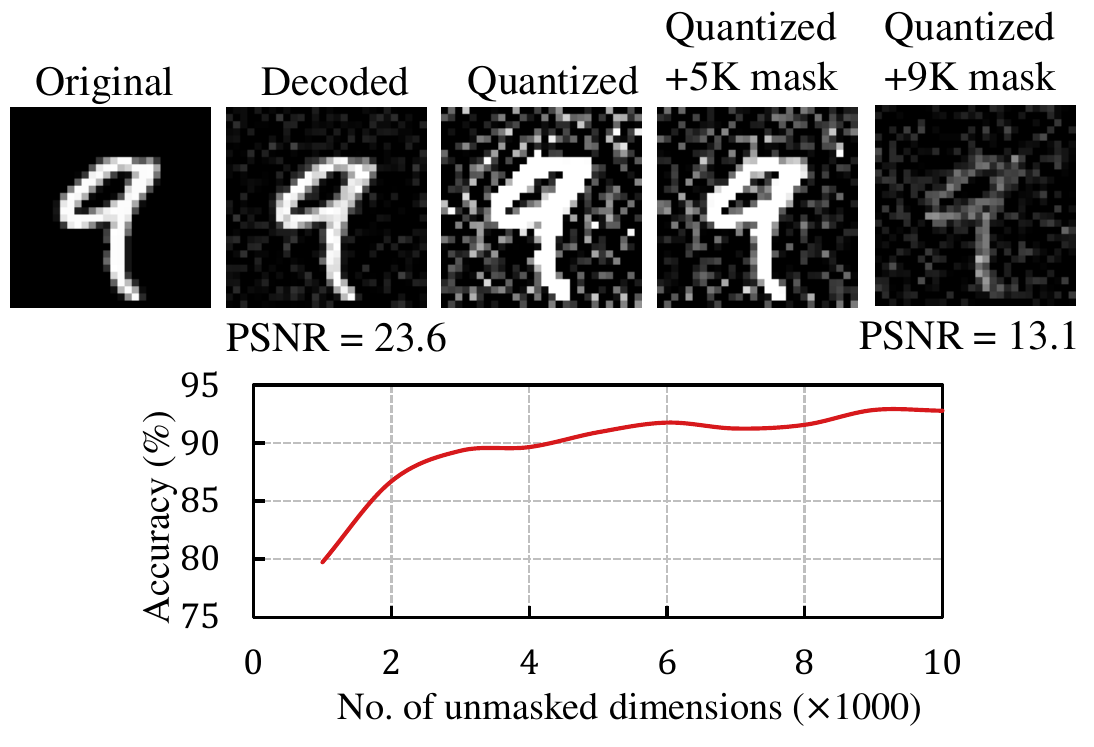}\vspace{-0.0cm}\caption{Impact of inference quantization and dimension masking on PSNR and accuracy.}\label{fig:inference}\vspace{-0.0cm}
  \end{figure}

Therefore, instead of sending the raw data, we propose to perform the light-weight encoding part on the edge and quantize the encoded vector before offloading to the remote host.
We call it \textit{inference quantization} do distinguish between encoding quantization, as inference quantization targets a full-precision model.
In addition, we also nullify a specific portion of encoded dimensions, i.e., mask out them to zero, to further obfuscate the information.
Remember that our technique does not need to modify or access to the trained model.

Fig. \ref{fig:inference} shows the impact of inference 1-bit quantization on the speech recognition model.
When only the offloaded information (i.e., query hypervector with 10,000 dimensions) is quantized, the prediction accuracy is 92.8\%, which is merely 0.5\% lower than the full-precision baseline.
By masking out 5,000 dimensions, the accuracy is still above 91\%, while the reconstructed image becomes blurry.
While the reconstructed image (from a typical encoded hypervector) has a PSNR of 23.6 dB, in our technique, it shrinks to 13.1.

\subsection{Hardware Optimization}\label{subsec:hardware}

\begin{figure}[t]
  \centering
  \includegraphics[width=0.49\textwidth]{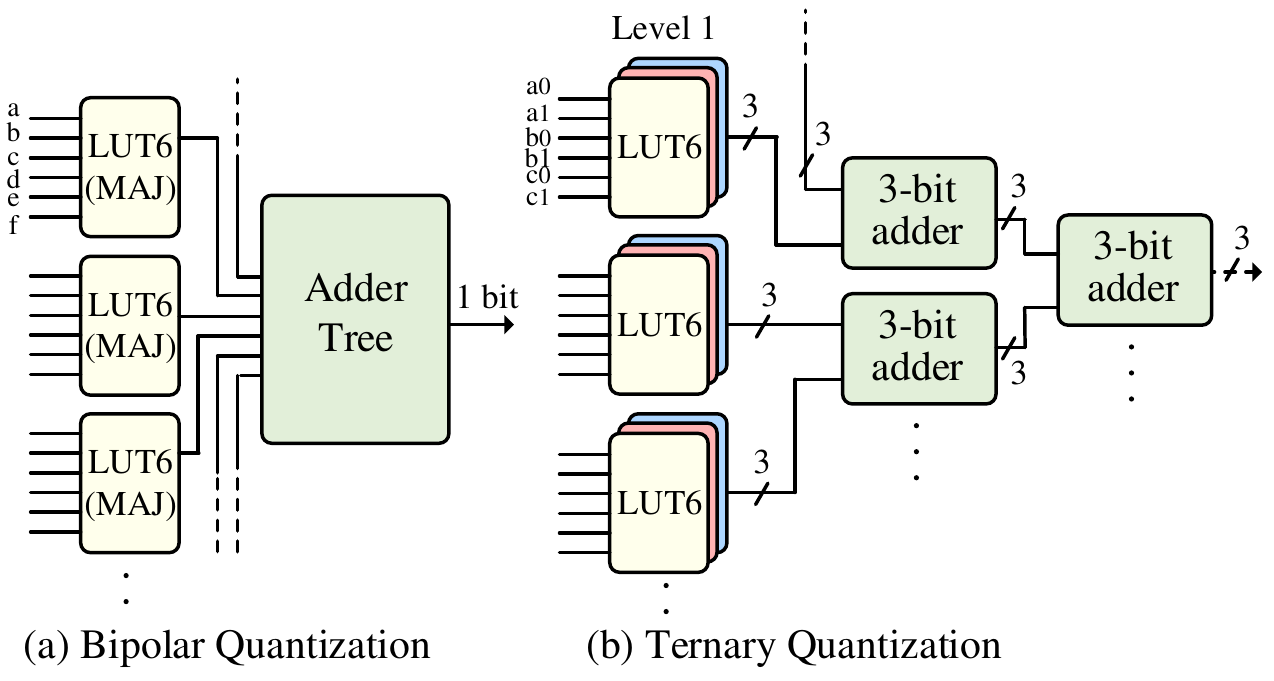}\vspace{-0.0cm}\caption{Principal blocks of FPGA implementation.}\label{fig:arch}\vspace{-0.0cm}
  \end{figure}

\begin{figure*}[t]
  \centering
  \includegraphics[width=1.0\textwidth]{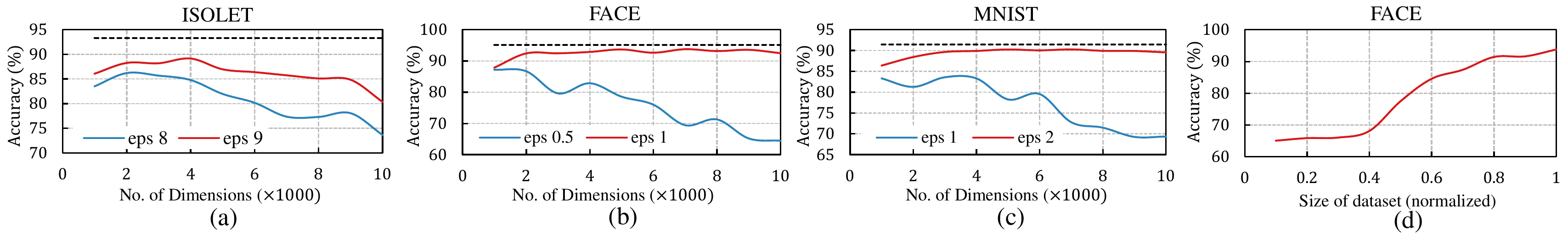}\vspace{-0.0cm}
  \caption{Investigating the optimal $\varepsilon$, dimensions and impact of data size in the benchmark models.}
  \label{fig:training} \vspace{-0.0cm}
\end{figure*}

The simple bit-level operations involved in the proposed techniques and dimension-wise parallelism of the computation makes FPGA a highly efficient platform to accelerate privacy-aware HD computing \cite{imani2019sparsehd, salamat2019f5}.
We devise efficient implementations to further improve the performance and power.
We adopt the encoding of Equation \eqref{eq:encb} as it provides better optimization opportunity.

For the 1-bit bipolar quantization, a basic approach is adding up all bits of the same dimension, followed by a final sign/threshold operation.
This is equivalent to a \textit{majority} operation between `$-1$'s and `$+$1's.
Note that we can represent $-$1 by 0, and $+$1 by 1 in hardware, as it does not change the logic behind.
We shrink this majority by approximating it as partial majorities.
As shown by Fig. \ref{fig:arch}(a), we use 6-input look-up tables (LUT-6) to obtain the majority of every six bits (out of $d_{iv}$ bits), which are binary elements making a certain dimension.
In the case an LUT has equal number of 0 and 1 inputs, it breaks the tie randomly (predetermined)
We can repeat this till $\log d_{iv}$ stages but that would degrade accuracy. Thus, we use majority LUTs only in the first stage, so the next stages are typical adder-tree \cite{imani2019sparsehd}.
This approach is not exact, however, in practice it imposes $< 1\%$ accuracy loss due to inherent error tolerance of HD, especially we use majority LUTs only in the first stage, so the next stages are typical adder-tree \cite{imani2019sparsehd}.
Total number of LUT-6s will be:

\begin{equation}\label{eq:popcount}
n_{\text{LUT6}} =  \frac{d_{iv}}{6} + \frac{1}{6} ( \sum_{i=1}^{\log d_{iv}}{\frac{d_{iv}}{3 } \times \frac{i}{2^{i-1}} ) \simeq \frac{7}{18} d_{iv}} 
\end{equation}

which is $70.8\%$ less than $\frac{4}{3}d_{iv}$ required in the exact adder-tree implementation.

For the ternary quantization, we first note that each dimension can be $\{0, \pm1\}$, so requires two bits.
The minimum (maximum) of adding three dimensions is therefore $-$3 ($+$3), which requires three bits, while typical addition of three 2-bit values requires four bits.
Thus, as shown in Fig. \ref{fig:arch}(b), we can pass numbers (dimensions) $\overline{a_1a_0}$, $\overline{b_1b_0}$ and $\overline{c_1c_0}$ to three LUT-6 to produce the 3-bit output.
Instead of using an exact adder-tree to sum up the resultant $\frac{d_{iv}}{3}$ three-bits, we use saturated adder-tree where the intermediate adders maintain a bit-width of three through truncating the least-significant bit of output.
In a similar fashion to Equation \eqref{eq:popcount}, we can show that this technique uses $\simeq 2d_{iv}$ LUT-6, saving 33.3\% compared to $\simeq 3d_{iv}$ in the case of using exact adder-tree to sum up $d_{iv}$ ternary values.

\section{Results} \label{sec:res}

\subsection{Differentially Private Training}
We evaluate the privacy metrics of the proposed techniques by training three models on different categories: the same speech recognition dataset (ISOLET) \cite{Isolet} we used within the paper, the MNIST handwritten digits dataset, and Caltech web faces dataset (FACE) \cite{griffin2007caltech}.
The goal of training evaluation is to find out the minimum $\varepsilon$ with affordable impact on accuracy.
Similar to \cite{abadi2016deep}, we set the $\delta$ parameter of the privacy to $10^{-5}$ (which is reasonable especially the size of our datasets are smaller than $10^{5}$).
Accordingly, for a particular $\varepsilon$, we can obtain the $\sigma$ factor of the required Gaussian noise (see Equation \eqref{eq:dif2}) from $\delta \geq \frac{4}{5} e^{-\frac{(\sigma \varepsilon)^2}{2}}$ \cite{abadi2016deep}.
We iterate over different values of $\varepsilon$ to find the minimum while the prediction accuracy remains acceptable.

Fig. \ref{fig:training}(a)--(c) shows the obtained $\varepsilon$ for each training model and corresponding accuracy.
For instance, for the FACE model (Fig. \ref{fig:training}(b)), $\varepsilon = 1$ (labeled by \textit{eps1}) gives an accuracy within 1.4\% of the non-private full-precision model.
Shown by the same figure, slightly reducing $\varepsilon$ to $0.5$ causes significant accuracy loss.
This figure also reveals where the minimum $\varepsilon$ is obtained.
For each $\varepsilon$, using the proposed pruning and ternary quantization, we reduce the dimension to decrease the sensitivity.
At each dimension, we inject a Gaussian noise with standard deviation of $\Delta f \cdot \sigma$ with $\sigma$ obtainable from $\delta = 10^{-5} = \frac{4}{5} e^{-\frac{(\sigma \varepsilon)^2}{2}}$, which is $\sim$4.75 for a demanded $\varepsilon = 1$.
$\Delta f$ of different quantization schemes and dimensions is already discussed and shown by Fig. \ref{fig:quant-sen}.
When the model has large number of dimensions, its primary accuracy is better, but on the other hand has higher sensitivity ($\propto \sqrt{\mathcal{D}_{hv}}$).
Thus, there is a trade-off between dimension reduction to decrease sensitivity (hence, noise) and inherent accuracy degradation associated with dimension reduction itself.
For FACE model, we see that optimal number of dimension to yield the minimum $\epsilon$ is 7,000.
It should be noted that although there is no prior work on HD privacy (and few works on DNN training privacy) for a head-to-head comparison, we could obtain a single digit $\varepsilon = 2$ for the MNIST dataset with $\sim$1\% accuracy loss (with 5,000 ternary dimensions), which is comparable to the differentially private DNN training over the MNIST in \cite{abadi2016deep} that achieved the same $\varepsilon$ with $\sim 4\%$ accuracy loss.
In addition, differentially private DNN training requires very large number of training epochs where the per-epoch training time also increases (e.g., by $4.5\times$ in \cite{abadi2016deep}) while we readily apply the noise after building up all class hypervectors.
We also do not retrain the noisy model as it violates the concept of differential privacy.

Fig. \ref{fig:training}(d) shows the impact of training data size on the accuracy of the FACE differentially private model.
Obviously, increasing the number of training inputs enhances the model accuracy.
This due to the fact that, because of quantization of encoded hypervectors, the class vectors made by their bundling have smaller values.
Thus, the magnitude of induced noise becomes comparable to the class values.
As more data is trained, the variance of class dimensions also increases, which can better bury the same amount of noise.
This can be considered a vital insight in privacy-preserved HD training.

\subsection{Privacy-Aware Inference}

\begin{figure}[t]
  \centering
  \includegraphics[width=0.50\textwidth]{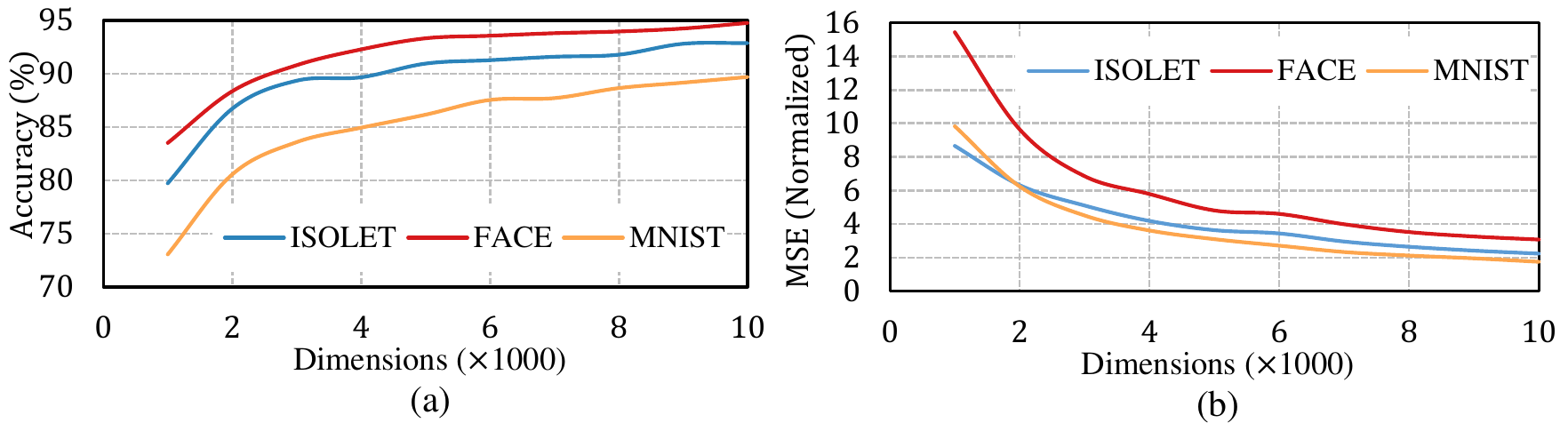}\vspace{-0.0cm}\caption{Impact of inference quantization (left) and dimension masking on accuracy and MSE.}\label{fig:res_inference}\vspace{-0.0cm}
  \end{figure}
  
Here we show a similar result of Fig. \ref{fig:inference} on HD models trained on different datasets.
Fig. \ref{fig:res_inference}(a) shows the impact of bipolar quantization of encoding hypervectors on the prediction accuracy.
As discussed in Section \ref{subsec:inference}, here we merely quantize the encoded hypervectors (to be offloaded to cloud for inference) while the class hypervectors remain intact.
Without pruning the dimensions, the accuracy of ISOLET, FACE, and MNIST degrades by 0.85\% on average, while the mean squared error of the reconstructed input increases by $2.36\times$, compared to the data reconstructed (decoded) from conventional encoding.
Since the dataset of ISOLET and FACE are extracted features (rather than raw data), we cannot visualize them, but from Fig. \ref{fig:res_inference}(b) we can observe that ISOLET gives a similar MSE error to MNIST (for which the visualized data can be seen in Fig. \ref{fig:inference}) while the FACE dataset leads to even higher errors.

In conjunction with quantizing the offloaded inference, as discussed before, we can also prune some of the encoded dimensions to further obfuscate the information.
We can see that in the ISOLET and FACE models, discarding up to 6,000 dimensions leads to a minor accuracy degradation while the increase of their information loss (i.e., increased MSE) is considerable.
In the case of MNIST, however, accuracy loss is abrupt and does not allow for large pruning.
However, even pruning 1,000 of its dimensions (together with quantization) reduces the PSNR to $\sim$15, meaning that reconstruction of our encoding is highly lossy.

\subsection{FPGA Implementation}
We implemented the HD inference using the proposed encoding with the optimization detailed in Section \ref{subsec:hardware}.
We implemented a pipelined architecture with building blocks shown in Fig. \ref{fig:arch}(a) as in the inference we only used binary (bipolar) quantization.
We used a hand-crafted design in Verilog HDL with Xilinx primitives to enable efficient implementation of the cascaded LUT chains.
Except the proposed approximate adders, the rest of implementation follows an architecture similar to \cite{imani2019sparsehd}.
Table \ref{tab:fpga} compares the results of Prive-HD on Xilinx Kintex-7 FPGA KC705 Evaluation Kit, versus software implementation on Raspberry Pi 3 embedded processor and NVIDIA GeForce GTX 1080 Ti GPU.
Throughout denotes number of inputs processed per second, and energy indicates energy (in Joule) of processing a single input.
All benchmarks have have the same number of dimensions in different platforms.
For FPGA, we assumed that all data resides in the off-chip DRAM, otherwise the latency will be affected but throughout remains intact as off-chip latency is eliminated in the computation pipeline.
Thanks to the massive bit-level parallelism of FPGA with relatively low power consumption ($\sim$7W obtained via Xilinx Power Estimator, compared to $3$W of Raspberry Pi obtained by Hioki 3334 power meter, and $120$W of GPU obtained through NVIDIA system management interface), the average inference throughput of Prive-HD is $105$,$067\times$ and $15.8\times$ of Raspberry Pi and GPU, respectively.
Prive-HD improves the energy by $52$,$896\times$ and $288\times$ compared to Raspberry Pi and GPU, respectively.

\begin{table}[t]
\caption{Comparing the Prive-HD on FPGA versus Raspberry Pi and GPU}
\resizebox{1.00\columnwidth}{!}{
\Huge
\begin{tabular}{|l|c|c|c|c|c|c|}
\hline
       & \multicolumn{2}{c|}{Raspberry Pi} & \multicolumn{2}{c|}{GPU} & \multicolumn{2}{c|}{Prive-HD (FPGA)} \\ \hline
       & Throughput        & Energy        & Throughput   & Energy    & Throughput         & Energy          \\ \hline
ISOLET & $19.8$              & $0.155$         & $135,300$      & $8.9\times10^{-4}$   & $2,500,000$          & $2.7\times10^{-6}$         \\ \hline
FACE   & $11.9$              & $0.266$         & $104,079$      & $1.2\times10^{-3}$   & $694,444$            & $4.7\times10^{-6}$         \\ \hline
MNIST  & $23.9$              & $0.129$         & $140,550$      & $8.5\times10^{-4}$   & $3,125,000$          & $3.0\times10^{-6}$         \\ \hline
\end{tabular}

}
\vspace{-0.0cm}
\label{tab:fpga}
\end{table}

\section{Conclusion}
In this paper, we disclosed the privacy breach of hyperdimensional computing and presented a privacy-preserving training scheme by quantizing the encoded hypervectors involved in training, as well as reducing their dimensionality, which together enable employing differential privacy by relieving the required amount of noise.
We also showed that we can leverage the same quantization approach in conjunction with nullifying particular elements of encoded hypervectors to obfuscate the information transferred for untrustworthy cloud (or link) inference.
We also proposed hardware optimization for efficient implementation of the quantization schemes by essentially using approximate cascaded majority operations.
Our training technique could address the discussed challenges of HD privacy and achieved single-digit privacy metric.
Our proposed inference, which can be readily employed in a trained HD model, could reduce the PSNR of an image dataset to below
15 dB with affordable impact on accuracy.
Eventually, we implemented the proposed encoding on an FPGA platform which achieved $15.8\times$ speed-up and $288.8\times$ energy efficiency over an optimized GPU implementation.

\section*{Acknowledgements}
This work was supported in part by CRISP, one of six centers in JUMP, an SRC program sponsored by DARPA, in part by SRC-Global Research Collaboration grant, and also NSF grants \#1527034, \#1730158, \#1826967, and \#1911095.

\bibliographystyle{IEEEtran}
\bibliography{references}

\end{document}